\documentclass[10pt]{asme2ej}

\usepackage{graphicx}
\usepackage{fancyhdr}
\usepackage{lastpage}
\title{NAVARO II, \goodbreak a novel scissor-based planar parallel robot\footnote{The original version of this paper has been accepted for presentation at the ASME 2018 International Design Engineering Technical Conferences \& Computers and Information in Engineering Conference, DETC2018-85949,  
August 26--29, 2018, Qu\'ebec, Qc, Canada}}

\author{Damien Chablat
    \affiliation{
	CNRS, Laboratoire des Sciences du Num\'erique de Nantes\\
	UMR CNRS 6004, 1 rue de la No\"e, \\
	44321 Nantes\\
    Email: damien.chablat@cnrs.fr
    }	
}

\author{Luc Rolland
    \affiliation{School of Computing, Engineering and Physical Sciences\\ University of the West of Scotland, \\
		Paisley, Scotland, UK\\
     Email: luc.rolland@uws.ac.uk
    }
}

\begin{document}
\maketitle    
\begin{abstract}
{\it This article presents a new variable actuation mechanism based on the 3-RPR parallel robot. This mechanism is an evolution of the NaVARo robot, a 3-RRR parallel robot, for which the second revolute joint of the three legs is replaced by a scissor to obtain a larger working space and avoid the use of parallelograms to operate the second revolute joint. To obtain better spatial rigidity, the leg mechanism is constructed by placing the scissors in an orthogonal plane to the plane of the manipulator displacement (3-RRR or even the 3-RPR). This geometric property brings the significant consequence of allowing the scissors to directly substitute the prismatic chains in the 3-RPR and enjoy the same kinematics advantages for the overall robots as only one solution to the inverse kinematic model. From the Jacobian expression, surfaces of singularity can be calculated and presented in a compact form. The singularity equations are presented for a robot with a similar  base and mobile platform. The properties of the scissors are then determined to have a prescribed regular workspace.}
\end{abstract}

\section{INTRODUCTION}
A major drawback of serial and parallel mechanisms is the inhomogeneity of kinetostatic performance in their workspace. For example, dexterity, accuracy and stiffness are generally poor in the vicinity of the singularities that may appear within the working space boundaries of these mechanisms. For parallel manipulators, their inverse kinematics problem often has several solutions, which can be considered as ``working mode'' \cite{Chablat:1998}. However, it is difficult to achieve a large workspace without any singularity for any given working mode. Therefore, it is necessary to plan a trajectory change for the working mode to avoid parallel singularities \cite{Chablat:1998bis,Wenger:2000}. In such a case, the initial trajectory would not be followed.

One solution to this problem is to introduce actuation redundancy, which could involve force control algorithms \cite{AlbaGomez:2005}. Another approach is to use the concept of joint coupling as proposed by ~\cite{Theingin:2007}. Moreover, in each leg where actuation reduncancy is implemented, it is possible to select the articulation which will be actuated in relation to the end-effector pose, ~\cite{Arakelian:2007}.

Practically, a first variable actuation mechanism (VAM) was introduced in 2008 \cite{Novona:2008}, called NaVARo for Nantes Variable Actuation Robot. This mechanism has eight actuation modes and is based on 3-RRR parallel topology, whereby the first or second revolute joint can be actuated. As this mechanism has eight solutions to the inverse kinematic model, the singularity determination and separation according to the current working mode is very difficult problem algebraically \cite{Bonev}. In addition, the volume swept by the robot's kinematics chain is large but the parallelograms reduce the workspace. A control framework has been developed to drive the robot prototype \cite{Chablat:2016}. The main problem comes from the localization of the position sensor on the motor, which is not always connected to the robot base. Additional sensors may separate assembly modes, but a slight slip in the couplings may disturb the location of the mobile platform. Another problem is the compliance errors according to the  actuation modes and the posture of the robot when the forces and torques applied on the mobile platform are not in the plane \cite{Klimchik:2018}.

The aforementioned solutions add between one and three actuators which require the hardware addition such as controller outputs, PWM amplifiers and cables which increase material requirements, increasing the complexity and capital costs. The proposed solution will virtually comprise six actuators but only connects three simultaneously through clutches driven by a digital output. As a result, the robot controller will only drive three actuators. The proposed solution will limit the driven actuators to three actuators including the PWM amplifiers and related analog elements thereby keeping the analog hardware to a minimum comparable to the original 3-RPR. The added requirements will be six clutches and their related digital hardware. Three encoders will determine the angular position of the fixed base revolute joints and three more will be fitted on the actuating screws to determine the displacement of the scissors. The scissor actuating screws will be selected reversible by either selecting the appropriate pitch or using ball screws allowing higher accelerations. This concept will be utilized and examined in this article.

The aim of this article is to propose a new mechanism, based on the 3-RPR parallel topology for which singularities are easier to calculate then the 3-RRR for all actuation modes. The possibility to change actuation mode will allow singularity avoidance. Among the other advantages, the actuated scissors provide linear displacements with less encumbrance then with prismatic or linear actuators. Finally, the perpendicularly located scissors offer increased rigidity in the third dimension perpendicular to the planar robot displacement plane. 

The outline of this article is as follows. The first section presents the architecture of the NaVARo II robot with its eight actuation modes. The second section examines kinematics modeling and presents the algebraic equations of parallel singularities as well as the limits of the workspace. The displacement ranges of the scissor mechanisms are determined to include a regular workspace inside. Depending on workspace limits, the singular surfaces are reduced to present only the significant singularities, the ones within the workspace.
\section{Mechanism architecture of the NaVARo II}
The VAM concept was examined in~\cite{Arakelian:2007,Theingin:2007}. They derived a VAM from the architecture of the 3-RPR planar parallel manipulator by actuating either the first revolute joint or the prismatic joint of its kinematics chains. The same concept was introduced in \cite{Novona:2008} based on the 3-RRR and a first prototype was reported in \cite{Mesrob}.

\begin{figure}
\begin{center}
\includegraphics[width=0.35\textwidth]{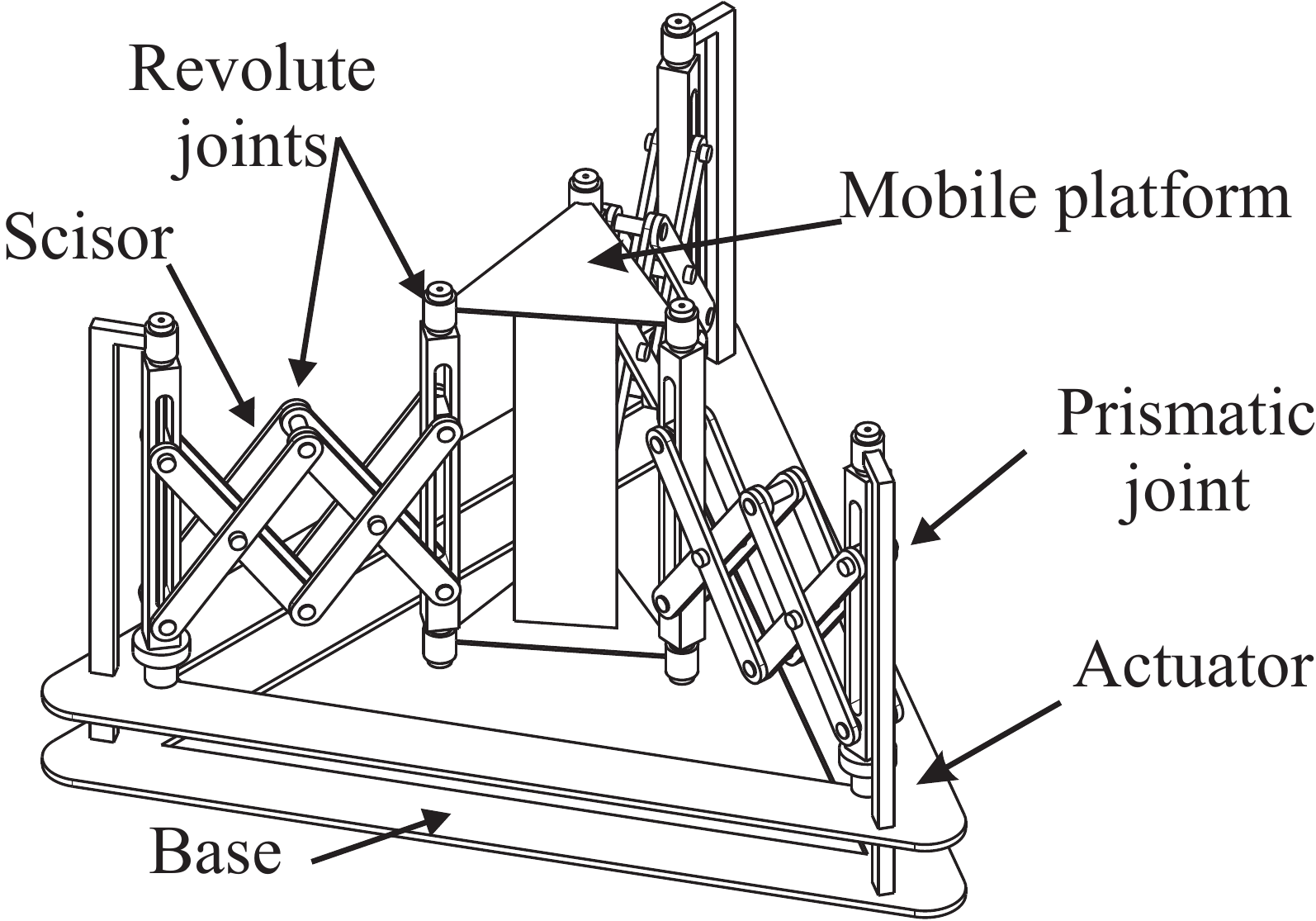}
\caption{Isometric view of NaVARo II}
\label{robot_iso}
\end{center}
\end{figure}

The newly proposed 3-RPR manipulator concept with variable actuation is shown in Fig.~\ref{robot_iso}. The use of scissors makes it possible to limit the space requirement and encumbrance during movements in contrast to previous designs where the prismatic actuators can bring signiciant encumbrance problems. Moreover, scissors are parallel mechanisms which improve rigidity. The number of scissors can be optimized according to the possible height of the mobile platform, the desired stiffness or the desired maximum length of the equivalent prismatic joint \cite{Takesue,Islam:2014,Rolland:2010}.

\begin{figure}
\begin{center}
\includegraphics[width=0.35\textwidth]{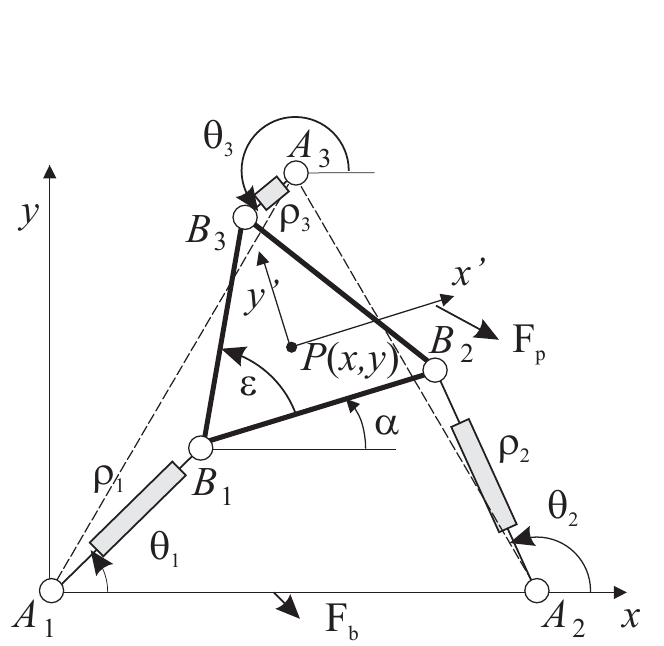}
\caption{3-RPR with variable actuation where the first revolute joint ($\theta_i$) or the prismatic joint ($\rho_i$) is actuated}
\label{3RPR}
\end{center}
\end{figure}

This mechanism can be represented on a projection like all 3-RPR mechanisms. The pose of the mobile platform is determined by the Cartesian coordinates ($x$, $y$) of the end-effector $P$ expressed in the basic frame $F_b$ and the angle $\alpha$, i.e. the angle between the reference frames $F_b$ and $F_p$ (Figure~\ref{3RPR}).   
To illustrate this manipulator, the following dimensions have been fixed, $A_1A_2=A_1A_3=A_2A_3=90$, $B_1B_2=B_1B_3=B_2B_3=30$, $\epsilon=\pi/3$ and $r_{min} \leq  \rho_i \leq r_{max}$ for $i=1, 2, 3$ where $\epsilon$ is the angle formed by $B_2, B_1, B_3$.

A new transmission system has been designed, developed and installed in each kinematics chain of the NaVARo II so that the robot controller can select one actuation mode or the other and easily switch between them through a dual clutch system. As for NaVARo I, it consists of one motor with a shaft where two clutches can provide motor torque to either the revolute joint or to the scissor mechanism for changing its length. The motor will drive 
either the revolute connection between the base and the kinematic chain or the specific prismatic joint displacing the scissor mechanism. 

\begin{figure}
\begin{center}
\includegraphics[width=0.7\textwidth]{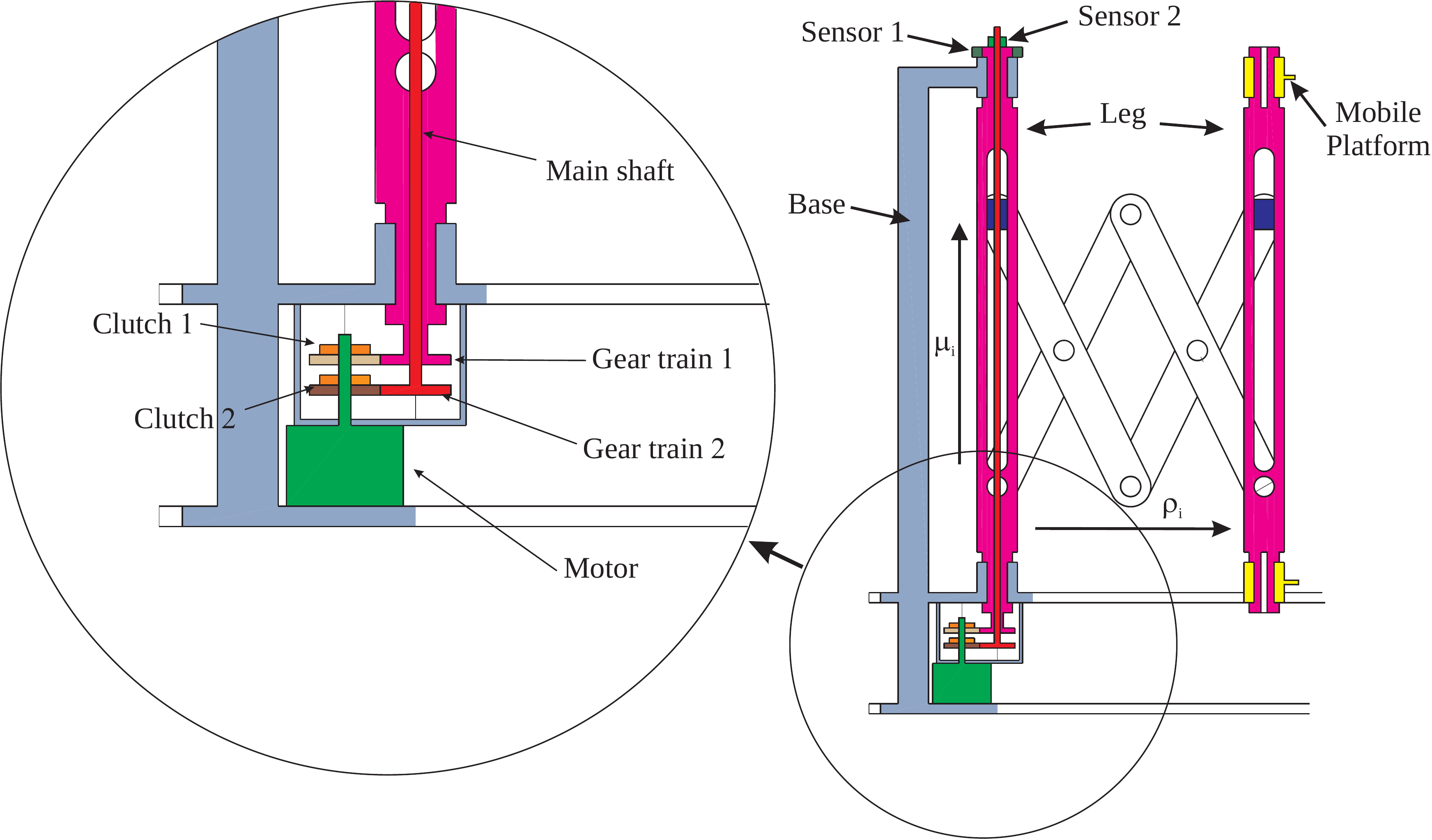}
\caption{The NAVARO II transmission system with two clutches and two gear trains for changing the actuation mode}
\label{actuation}
\end{center}
\end{figure}

Figure~\ref{actuation} shows an actuation diagram of the NaVARo II. This mechanical system contains: (i) an electric motor (green), (ii) a main shaft (red), (iii) a base (blue), (iv) the first axis of the kinematics chain (magenta), (v) a prismatic link shaft (yellow), and (vi) two electromechanical clutches (orange) which connects the motor axis to the main shaft or the first axis of the kinematics chain thanks to two gear trains. The gear ratios can be similar or defined according to the pitch of the ball screw to have similar maximum speed for the mobile platform when the actuation mode is changed. 

In each kinematics chain, through the appropriate clutches, the robot controller will select the first actuation mode by connecting the motor to the revolute joint or the second actuation mode by connecting the motor to the scissor. Note that the scissor operates with either a ball screw or machine screw where the pitch allows reversibility.

Two position sensors (preferably encoders) provide for the angular position of the leg relative to the base (sensor 1: $\theta_i$) and the angular position of the ball screw driving the scissor mechanism acting as the prismatic actuator (sensor 2: $\mu_i$). The values of the two sensors, combined with the joint limits, allow us to determine the current assembly mode of the robot. The length of the scissor $\rho_i$ is  computed thanks to the position of prismatic link shaft $\mu_i$ and the length $l$ of the scissors.

The actuation modes are slightly different from the NaVARo I. Each transmission system has four actuation schemes, that
are defined thereafter:
\begin{itemize}
\item 1. {\bf None of clutches 1 and 2 are active}. The main shaft can move freely in relation to the base. In this case, neither the pivot joint nor the prismatic joint is actuated. The kinematics chain can move freely, i. e. $\theta_i$ or $\rho_i$ are passive, $i = 1,2,3$. This mode allows the user to take and freely move the robot end-effector around the workspace. 
\item 2. {\bf Clutch 1 is active while clutch 2 is not}. The first kinematics chain axis (green) is driven by the rotation of the motor shaft. In this case, the angle $\theta_i$ is active while $\rho_i$ is passive, i = 1,2,3.
\item 3. {\bf Clutch 2 is active while clutch 1 is not}. The first kinematics chain joint is free but the rotation of the motor shaft leads to a slider displacement, which activates the scissor. In this case, the $\theta_i$ is passive and $\rho_i$ is active, $i = 1,2,3$.
\item 4. {\bf Both clutches 1 and 2 are active}. Both joints are are actuated by the same motors through some gear trains cause one displacement which would require synchronized rotation and translation. The end of the kinematics chain will make an almost spiral-like motion; however, the coupled leg constraints would limit the workspace and therefore will not be implemented.
\end{itemize}
The latter actuation mode differs from the NaVARo I. During path planning, only the second and third actuation modes will be implemented in the control algorithms and thereby examined in our study. Thus, NaVARo II has eight actuation modes, as shown in Table 1.  For example, the first actuation mode corresponds to the 3-\underline{R}PR mechanism, also referred to as the \underline{R}PR$_1$-\underline{R}PR$_2$-\underline{R}PR$_3$ mechanism, since the first revolute joint (located at point $A_i$) of its kinematics chain are actuated. Similarly, the eighth actuation mode corresponds to the 3-R\underline{P}R manipulator, also known as the R\underline{P}R$_1$-R\underline{P}R$_2$-R\underline{P}R$_3$ mechanism, since the prismatic joints of its kinematics chains are actuated. 
\begin{table}
  \begin{center}
\caption{The eight actuating modes of the 3-RRR VAM}
\begin{tabular}{|c|c|c|}
\hline 
\multicolumn{2}{|c|}{Actuating mode number} & active joints \\
\hline 
1&$\underline{R}PR_1$-$\underline{R}PR_2$-$\underline{R}PR_3$& 
$\theta_1$, $\theta_2$, $\theta_3$ \\
2&$\underline{R}PR_1$-$\underline{R}PR_2$-$R\underline{P}R_3$ &  
$\theta_1$, $\theta_2$, $\rho_3$ \\
3&$\underline{R}PR_1$-$R\underline{P}R_2$-$\underline{R}PR_3$ & 
$\theta_1$, $\rho_2$, $\theta_3$ \\
4&$R\underline{P}R_1$-$\underline{R}PR_2$-$\underline{R}PR_3$ &
 $\rho_1$, $\theta_2$, $\theta_3$ \\
5&$\underline{R}PR_1$-$R\underline{P}R_2$-$R\underline{P}R_3$ &
 $\theta_1$, $\rho_2$, $\rho_3$ \\
6&$R\underline{P}R_1$-$R\underline{P}R_2$-$\underline{R}PR_3$ &
 $\rho_1$, $\rho_2$, $\theta_3$ \\
7&$R\underline{P}R_1$-$\underline{R}PR_2$-$R\underline{P}R_3$ &
$\rho_1$, $\theta_2$, $\rho_3$ \\
8&$R\underline{P}R_1$-$R\underline{P}R_2$-$R\underline{P}R_3$ &
$\rho_1$, $\rho_2$, $\rho_3$ \\
\hline \hline
\end{tabular}
\label{tableau_robot}
  \end{center}
\end{table}
\section{Kinematic modeling of the NaVARo II manipulator}
In this section, we present the kinematic model that is commonly used to geometrically define the constraint equations, singular configurations, the workspace boundaries and surfaces that define the singularity loci.
\subsection{Kinematic modeling}
The velocity $\dot{\bf p}$ of point $P$ can be obtained in three different forms, depending on which kinematics chain is traversed~:
\begin{eqnarray}
\dot{\bf p}&=& \dot{\theta_i} {\bf E} ({\bf b}_i - {\bf a}_i) + \dot{\rho}_i \frac{{\bf b}_i - {\bf a}_i}{||{\bf b}_i - {\bf a}_i||}  + \dot{\alpha} {\bf E } ({\bf p} - {\bf b}_i)  \label{kinematics}
\end{eqnarray}
with matrix {\bf E} defined as
\begin{equation}
 {\bf E}=\left[\begin{array}{cc}
              0 & -1 \\
              1 &  0
             \end{array}
        \right]                
\end{equation}
Thus, ${\bf p}$, ${\bf b}_i$, ${\bf a}_i$ are the position vectors of points $P$, $A_i$ and $B_i$, respectively, and $\dot{\alpha}$ is the rate of angle $\alpha$.

To compute the kinematic model, we have to eliminate the idle joints $\theta_i$ or $\rho_i$ as a function of the actuation mode. We have to left multiply Eq.~\ref{kinematics} by ${\bf h}_i^T$ defined, for $\dot{\theta}_i$ as
\begin{equation}
{\bf h}_i=({\bf b}_i- {\bf a}_i)
\end{equation} 
and for $\dot{\rho}_i$ as
\begin{equation}
{\bf h}_i={\bf E} \frac{{\bf b}_i - {\bf a}_i}{||{\bf b}_i - {\bf a}_i||}.
\end{equation}

\begin{figure}
   \begin{minipage}[c]{.49\linewidth}
      \begin{center}
       \includegraphics[width=0.5\textwidth]{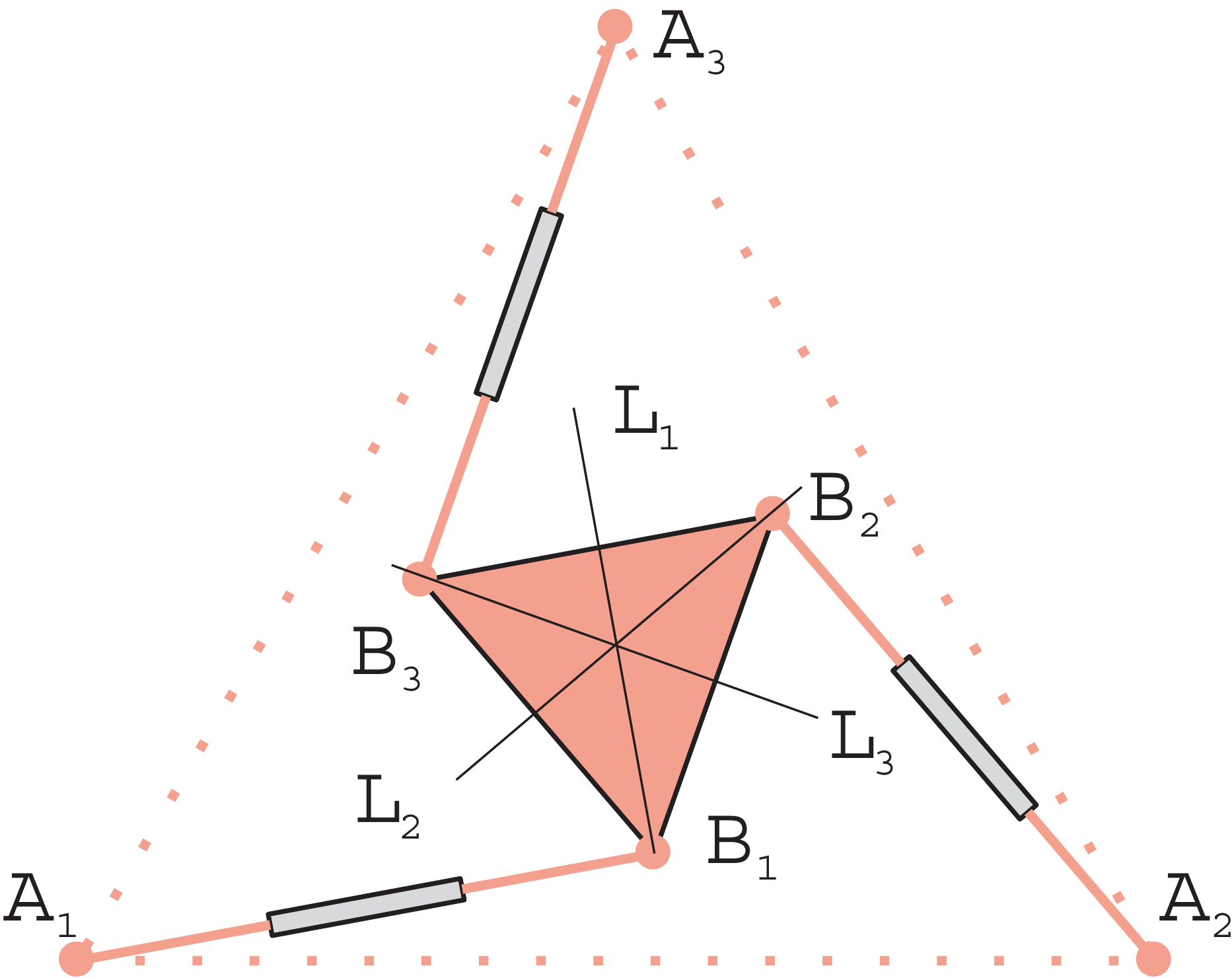}
       \caption{Example of singular configuration for the first actuation mode when the lines $L_1$, $L_2$ and $L_3$ intersect at one point}
       \label{Sing_RRR}
      \end{center}
   \end{minipage} \hfill
   \begin{minipage}[c]{.49\linewidth}
      \begin{center}
       \includegraphics[width=0.5\textwidth]{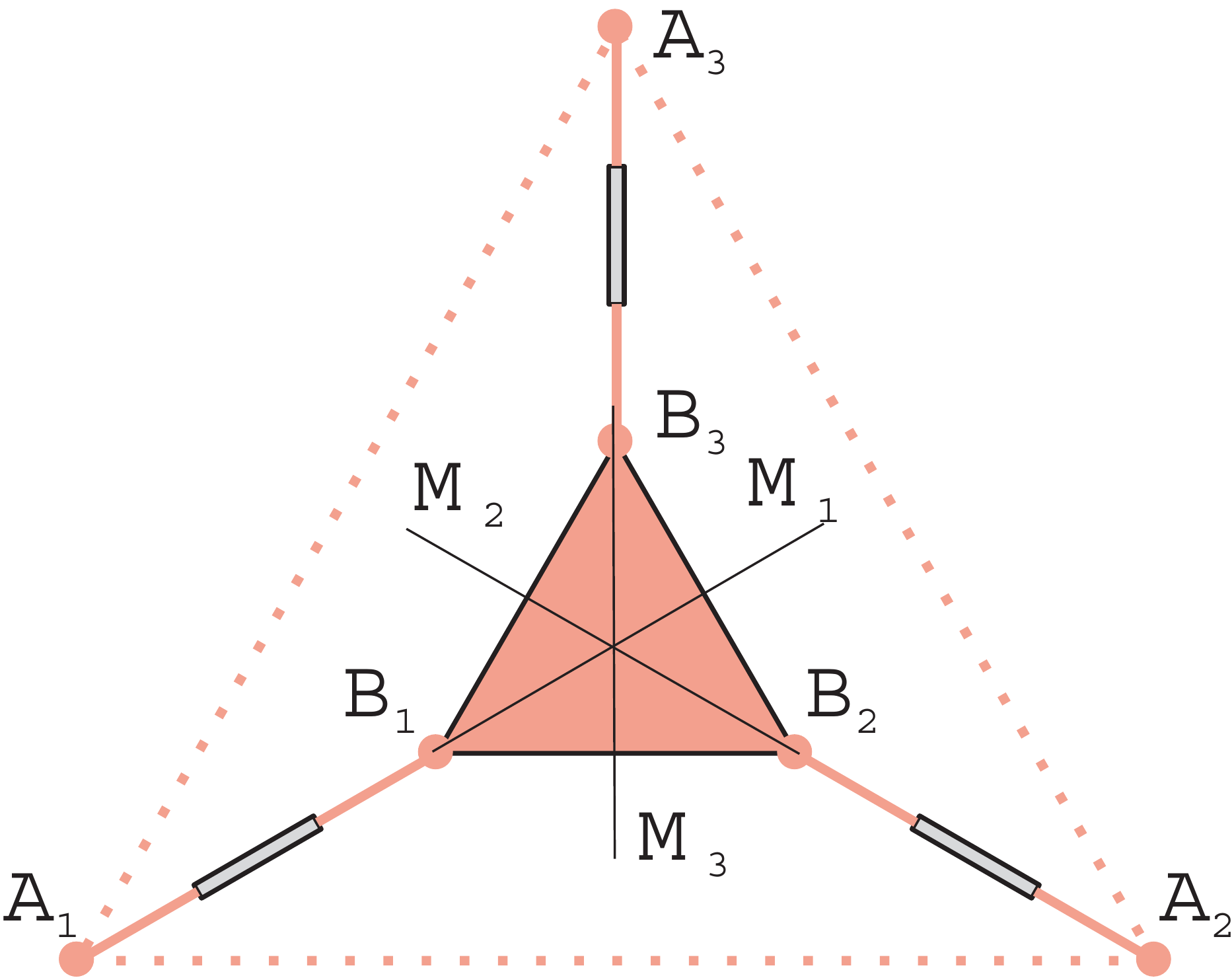}
       \caption{Example of singular configuration for the eighth actuation mode when the lines $M_1$, $M_2$ and $M_3$ intersect at one point}
       \label{Sing_PPP}
      \end{center}
   \end{minipage}
\end{figure}

The kinematic model of the VAM can now be expressed in vector form, namely,
\begin{equation}
{\bf A t} = {\bf B \dot{q}} \quad {\rm with} \quad {\bf t} = [\dot{\bf p}^T~~\dot{\alpha}]^T \quad {\rm and} \quad \dot{\bf q} = [\dot{q}_1~\dot{q}_2~\dot{q}_3]^T
\end{equation}

with ${\dot {\bf q}}_i$ thus being the vector of actuated joint rates where $\dot {q}_i = \dot{\theta}_i$ when the first revolute joint is driven and $\dot{q}_i = \dot{\rho}_i$ when the prismatic joint is driven, for $i=1,2,3$. {\bf A} and {\bf B} are respectively, the direct and the inverse Jacobian matrices of the mechanism, defined as
\begin{equation}
{\bf A}= \left[ {\begin{array}{*{20}{c}}
{{\bf{h}}_1^T}&-{{\bf{h}}_1^T{\bf{E}}({\bf{p}} - {\bf{b}_1})}\\
{{\bf{h}}_2^T}&-{{\bf{h}}_2^T{\bf{E}}({\bf{p}} - {\bf{b}_2})}\\
{{\bf{h}}_3^T}&-{{\bf{h}}_3^T{\bf{E}}({\bf{p}} - {\bf{b}_3})}
\end{array}} \right] \quad
{\bf B}= {\rm diag}[\rho_1~\rho_2~\rho_3]
\end{equation}
The geometric conditions for parallel singularities are well known in the literature for the first and eighth actuation modes.
For the first actuation mode, it is when the lines $L_i$, normal to the axis $(A_iB_i)$ are intersecting  at one point, see Fig.~\ref{Sing_RRR}.

For the eighth actuation mode, it is when the lines $M_i$, passing through the axis $(A_iB_i)$ are intersecting at one point, see Fig.~\ref{Sing_PPP}. For the other modes, it is just necessary to consider either the $L_i$ or $M_i$ lines according to the actuated joints, i.e. $L_i$ when the $i^{th}$ revolute joint is actuated and $M_i$ when the $i^{th}$ prismatic joint is actuated.
\subsection{Constraint equations}
To maintain the robot symmetry, the end-effector position is located in the center of the mobile platform. The constraint equations for all actuation modes can be determined by traversing the closed loops of the mechanism. Equations~\ref{Eq_loop1}-\ref{Eq_loop2} describe the two closed loops and Eqs.~\ref{Eq_loop3}-\ref{Eq_loop4} define the position and orientation of the mobile platform.
\begin{eqnarray}
\rho_1 C_{\theta_1} +30 C_{\alpha} -\rho_2 C_{\theta_2} -90&=&0 \label{Eq_loop1}\\
\rho_1 S_{\theta_1} +30 S_{\alpha} -\rho_2  S_{\theta_2} &=&0 \\
\rho_1 C_{\theta_1} +15 (C_{\alpha} -\sqrt{3} S_{\alpha} -3)-\rho_3 C_{\theta_3} &=&0 \\
\rho_1 S_{\theta_1} +15 (S_{\alpha} +\sqrt{3} C_{\alpha} -3 \sqrt{3})-\rho_3 S_{\theta_3} &=&0 \label{Eq_loop2}\\
x-\rho_1 C_{\theta_1} -15 C_{\alpha} +5 \sqrt{3} S_{\alpha} &=&0 \label{Eq_loop3}\\
y-\rho_1 S_{\theta_1} -15 S_{\alpha} -5 \sqrt{3} C_{\alpha} &=&0 \label{Eq_loop4}
\end{eqnarray}
with
$C_{\alpha}= \cos(\alpha)$, $S_{\alpha}= \sin(\alpha)$, $C_{\theta_i}= \cos(\theta_i)$, $S_{\theta_i}= \sin(\theta_i)$ for $i=1, 2, 3$. To make these equations algebraic, we use a substitution of all trigonometric functions as well as the square root function with
\begin{eqnarray}
\sqrt{3}&=&S3 \quad {\rm and} \quad S3^2=3 \nonumber\\
\cos(\beta)&=&C_{\beta} \quad {\rm and} \quad \sin(\beta)=S_{\beta} \quad {\rm for~any~angles~} \beta.\nonumber
\end{eqnarray}
We obtain a system with eleven equations, four for loop closures, two for the position and orientation of the mobile platform, four for trigonometric functions and one for the square root function with 14 unknowns. This equation system is under-determined. In this case, the equation manipulation remains challenging and powerful algebraic tools must be used like the Siropa library implemented in Maple \cite{Siropa,Siropa_MMT}.
\subsection{Workspace boundaries}
\begin{figure}
\begin{center}
\includegraphics[height=5cm]{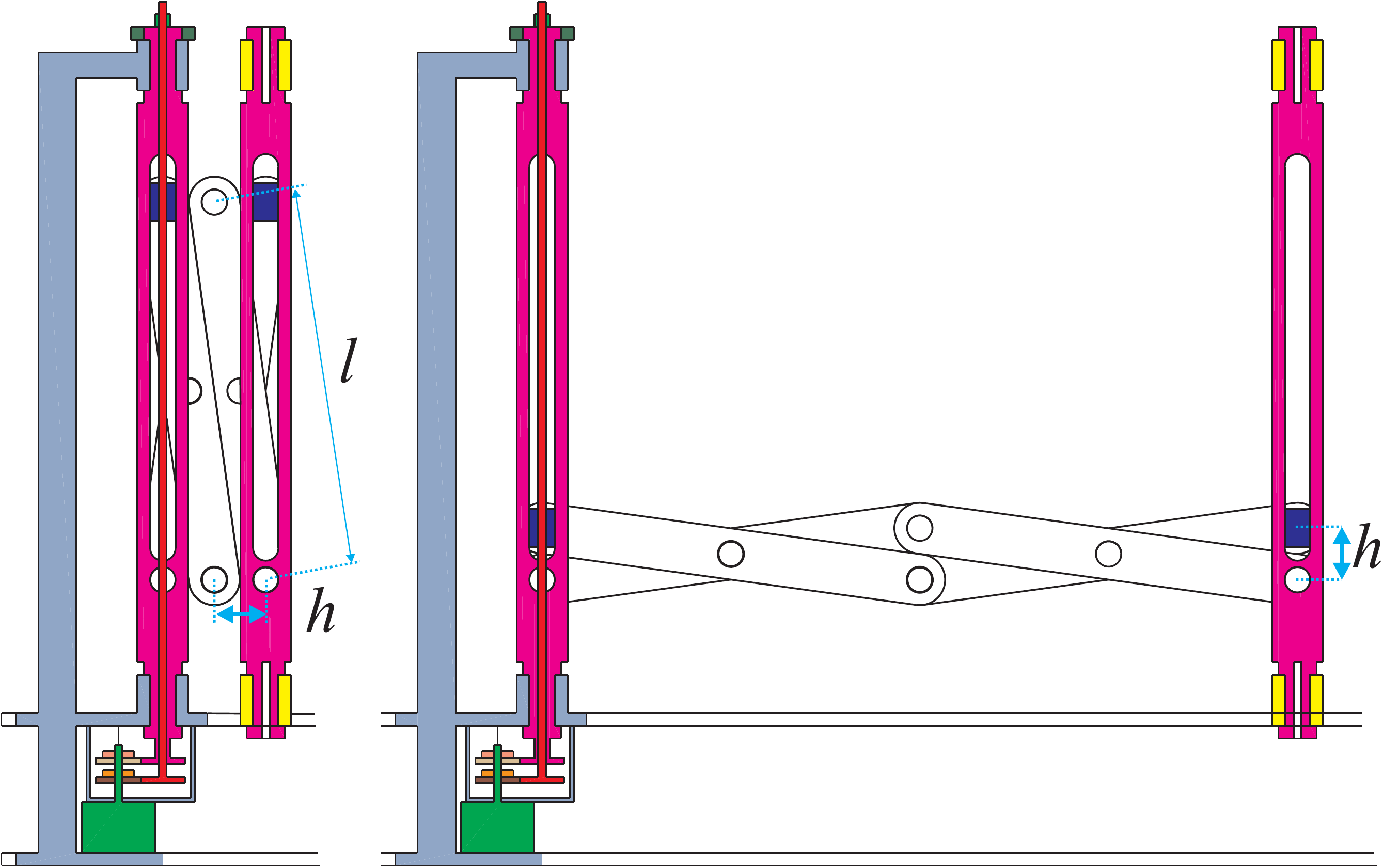}
\caption{Minimum and maximum lengths of the scissors}
\label{Ciseaux}
\end{center}
\end{figure}

If the revolute joints have no limits, the boundary of the workspace is given by the minimum and maximum extension of the scissor mechanisms as shown in Fig.~\ref{Ciseaux}. The minimum value of $\rho_i$ permits to avoid serial singular configuration where $\rho_i=0$.
Using the constraint equations and ranges limits of prismatic joints, ($r_{min} \leq  \rho_i \leq r_{max}$), which correspond to the scissor mechanisms, we find three surface equations that describe the boundary of the working space. These limits mean that there is no collision between the kinematic chains and the mobile platform. 
\begin{eqnarray}
10\,\sqrt {3}x\sin \left( \alpha \right) -10\,\sqrt {3}y\cos \left(\alpha \right) - r_{max}^2+x^2-30\,x\cos \left( \alpha
 \right) +y^2-30\,y\sin \left( \alpha \right) +300&=&0 \label{s1}\\
 \left( -10\,y\cos \left( \alpha \right) +10\,\sin \left( \alpha \right)  \left( x-90 \right)  \right) \sqrt {3}+ \left( 30\,x-2700
 \right) \cos \left( \alpha \right) +{x}^{2}+{y}^{2}+30\,y\sin \left(\alpha \right) - r_{max}^2-180\,x+8400&=&0 \label{s2}\\
 \left( 20\,y\cos \left( \alpha \right) + \left( -20\,x+900 \right) 
\sin \left( \alpha \right) -90\,y \right) \sqrt {3}+{x}^{2}+{y}^{2}-r_{max}^2-90\,x-2700\,\cos \left( \alpha \right) +8400 &=&0\label{s3}
\end{eqnarray}
\begin{figure}
\includegraphics[width=0.90\textwidth]{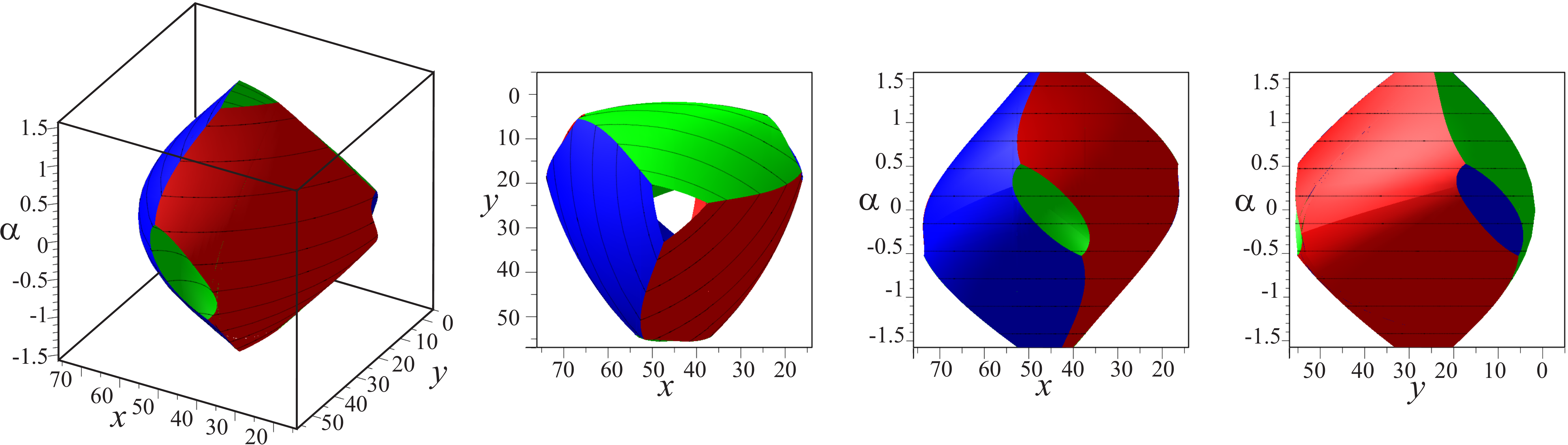}
\caption{Workspace of the NaVARo II in isometric view and three projections onto the planes ($xy$), ($x\alpha$) and ($y\alpha$)}
\label{workspace}
\end{figure}
For example, Fig.~\ref{workspace} depicts the boundaries of the workspace where $r_{min}=8$ and $r_{max}=59$. One function of the Siropa library displays surfaces that can be limited by inequality equations. The surfaces are shown in blue, red and green represent the minimum and maximum limits of leg one, two and three, respectively in Fig.~\ref{workspace}. The projections onto the ($xy$), ($x\alpha$) and ($y\alpha$) planes are used to estimate the main dimensions of the workspace. A cylindrical algebraic decomposition (CAD) can also be performed to have a partition of the workspace for each actuation mode \cite{Cbook:75}.
\subsection{Definition of a regular workspace}
\begin{figure}
\begin{center}
\includegraphics[height=4cm]{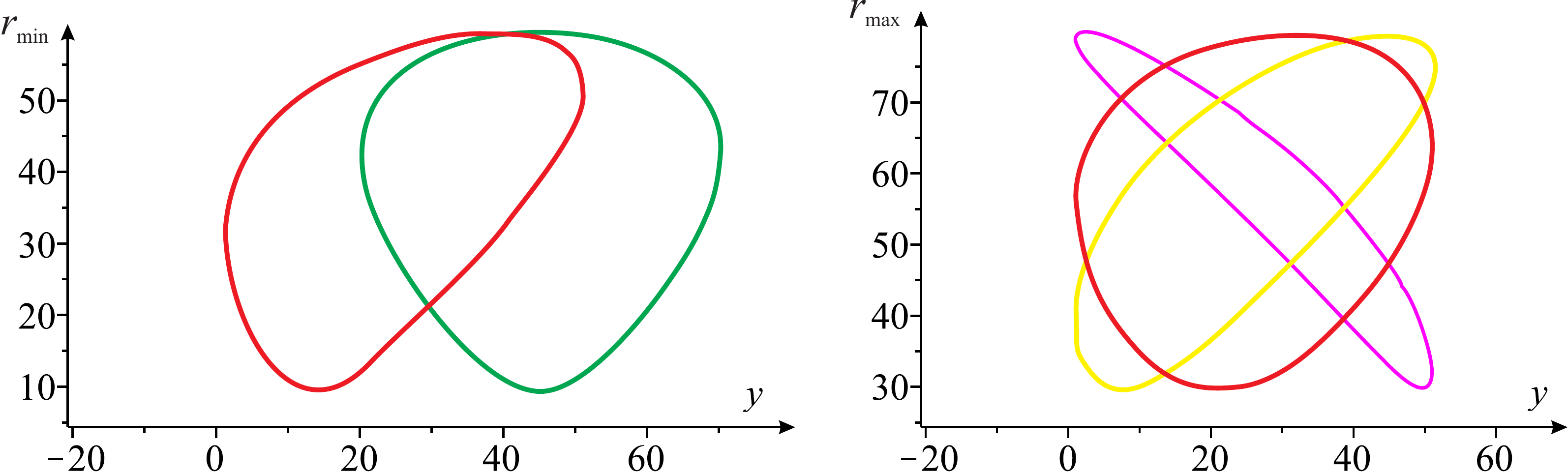}
\caption{Minimum and maximum lengths of the scissors for $r_{w}=25$}
\label{Couplage}
\end{center}
\end{figure}

Due to the selected symmetrical architecture, the simple regular workspace can be defined as a disc of radius $r$ whose centre is located at the barycentre of the triangle ($A_1A_2A_3$). For each position of the mobile platform, the orientation can vary over the following interval, $[\alpha_{min}~\alpha_{max}]$. As a numerical application, we choose the following parameters for our regular workspace, $r_{w}=25$, $\alpha_{min}= -\pi/2$ and $\alpha_{max}= \pi/2$. From Eqs.~\ref{s1}-\ref{s3}, we can observe that the minimum values of $r$ are reached for $\alpha=0$ and the maximal value of $r$ are reached for $\alpha=\alpha_{min}$ or $\alpha=\alpha_{max}$.  
\begin{eqnarray}
{\rm For~} \alpha&=&\pi/2: 10\,\sqrt {3}x+300- r_{max}^{2}+{x}^{2}+{y}^{2}-30\,y=0 \\
{\rm For~} \alpha&=&\pi/2:  \left( -900+10\,x \right) \sqrt {3}+8400+{x}^{2}+{y}^{2}+30\,y- r_{max}^{2}-180\,x = 0\\
{\rm For~} \alpha&=&\pi/2: \left( 900-20\,x-90\,y \right) \sqrt {3}+{x}^{2}+{y}^{2}- r_{max}^{2}-90\,x+8400=0 \\
{\rm For~} \alpha&=&0: 300-10\,\sqrt {3}y- r_{min}^{2}+{x}^{2}-30\,x+{y}^{2}=0 \\
{\rm For~} \alpha&=&0:-10\,\sqrt {3}y-150\,x+5700+{x}^{2}+{y}^{2}-r_{min}^{2}=0 \\
{\rm For~} \alpha&=&0:-70\,\sqrt {3}y+{x}^{2}+{y}^{2}-r_{min}^{2}-90\,x+5700=0 \\
{\rm For~} \alpha&=&-\pi/2:-10\,\sqrt {3}x+300- r_{max}^{2}+{x}^{2}+{y}^{2}+30\,y=0 \\
{\rm For~} \alpha&=&-\pi/2: \left( 900-10\,x \right) \sqrt {3}+8400+{x}^{2}+{y}^{2}-30\,y- r_{max}^{2}-180\,x=0 \\
{\rm For~} \alpha&=&-\pi/2: \left( -900+20\,x-90\,y \right) \sqrt {3}+{x}^{2}+{y}^{2}- r_{max}^{2}-90\,x+8400=0 
\end{eqnarray}
The $r$ range is defined when there is no intersection with the regular workspace. Finding intersections with the workspace boundary allows to determine the values of $r$. By expressing $x$ as a function of $y$, we can examine the number of intersections using the coupling curves shown in the figures~\ref{Couplage}. We obtain six equations which have the following shape for the intersection of the kinematic chain workspace which is fixed in $A_1$
\begin{eqnarray}
10\,\sqrt{3} \left( 45+\sqrt {-50+30\,\sqrt {3}y-y^2} \right) +300-r^2+ \left( 45+\sqrt {-50+30\,\sqrt {3}y-y^2} \right)^2+y^2-30\,y &=&0\\
10\,\sqrt{3} \left( 45-\sqrt {-50+30\,\sqrt {3}y-y^2} \right) +300-r^2+ \left( 45-\sqrt {-50+30\,\sqrt {3}y-y^2} \right)^2+y^2-30\,y &=&0
\label{eq:exemple}
\end{eqnarray}

To find the maximum or minimum value of $r$ on these curves, it is sufficient to study the sense of variation by deriving with respect to $y$. In our case study, we find $r_{min}= 9.64$ and  $r_{max}= 79.77$.

From Fig.~\ref{Ciseaux}, the length of each scissor bar and its height allow to define, for $n$ scissors, the minimum and maximum length, $r_{min}$ and $r_{max}$ respectively, as well as a relationship between height and length
\begin{eqnarray}
r_{min}&=& n h \\
r_{max}&=& n \sqrt{l^2 - h^2}\\
l&>&3h
\label{eq:r_ciseaux}
\end{eqnarray}

\begin{figure}
\begin{center}
\includegraphics[width=0.45\textwidth]{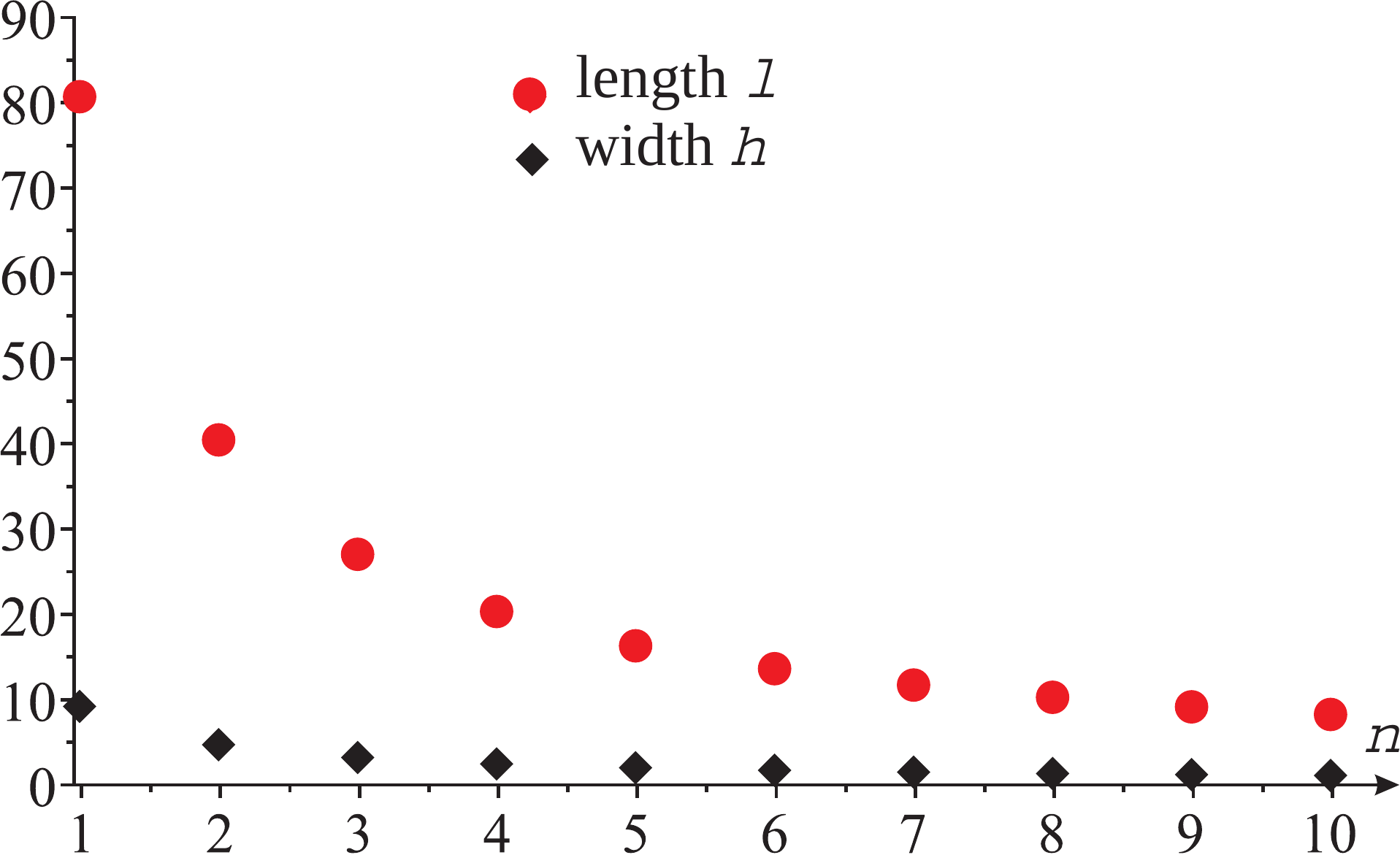}
\caption{Evolution of the width $h$ and length $l$ of the bars according to the number of scissors $n$}
\label{ciseaux_largeur_hauteur}
\end{center}
\end{figure}

We can make a numerical application from the previous results according to the number of scissors. 
\begin{equation}
h= 9 / n \quad
l= \sqrt{6481} / n
\label{eq:h_n}
\end{equation}
The graphics in Figure~\ref{ciseaux_largeur_hauteur} show the length $l$ and height $h$ of the scissors as a function of the $n$ number of scissors. The inequality constraint is always checked but the bar cross-sections decrease with $n$, so the rigidity of the structure is conjectured to decrease. A stiffness study is therefore necessary to choose the maximum number of scissors that can be used. This could lead to an interesting optimisation problem left for further research.
\section{Singular configurations}
From the constraint equations, it is possible to write the determinant of matrix {\bf A}. These determinants depend on the positions of the mobile platform, the passive joints and the active joints. Only an elimination method based on Groebner's basis can successfully obtain the singularity representation in the Cartesian workspace. Note that for only the first and eighth actuation modes, the determinant of {\bf A} is factorized to form two parallel planes. For the eighth actuation mode, an unrepresented singularity exists for $\alpha=\pi$. The equations of the singularities for the eight actuation modes are given in the Appendix. As there is only one working mode, the equations of these surfaces are simpler than for the NAVARO I robot for which it is not possible to simply describe these equations in a sufficiently compact format for publication.

\begin{figure}
   \begin{minipage}[c]{.49\linewidth}
      \begin{center}
      \includegraphics[width=0.6\textwidth]{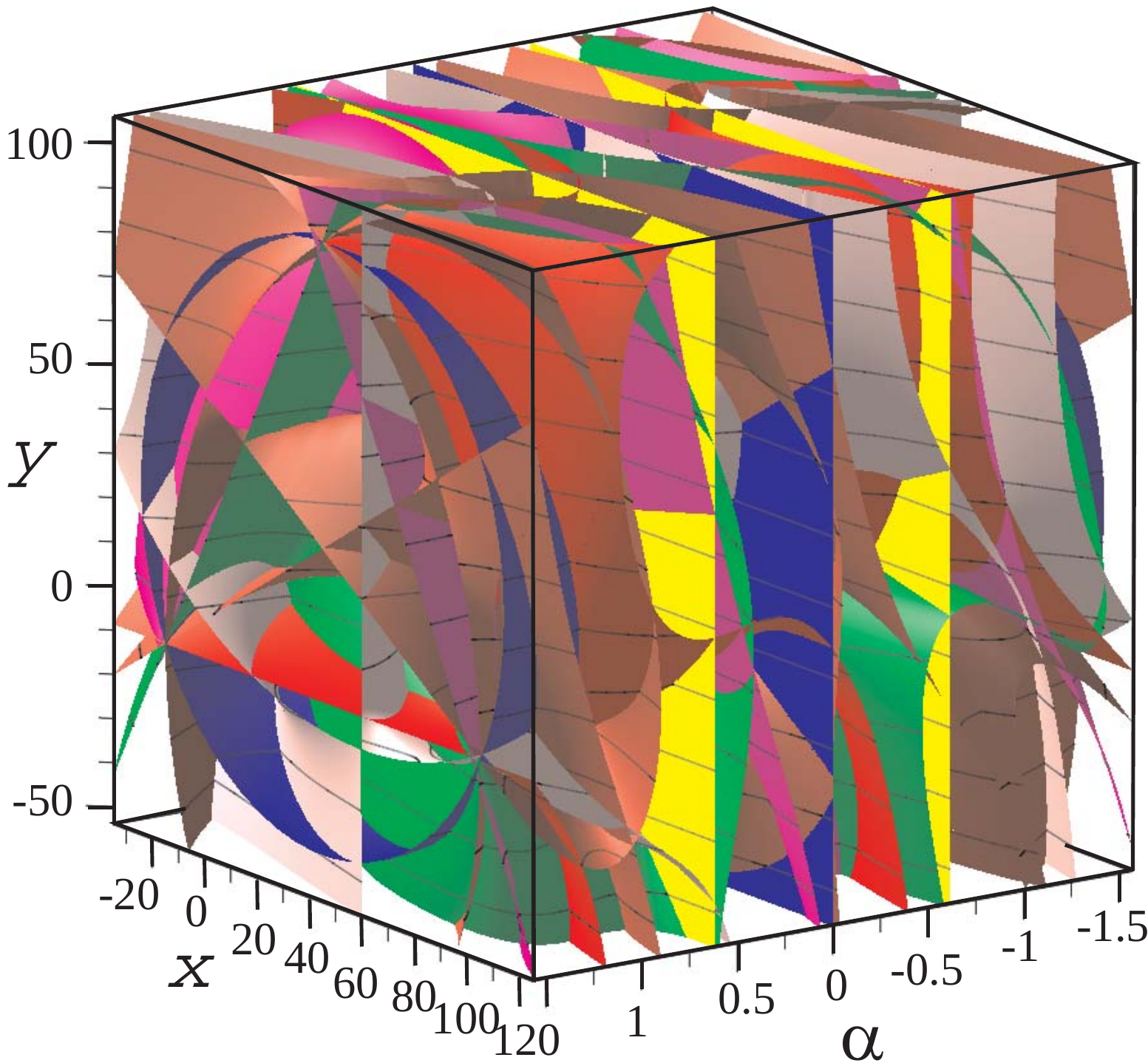}
      \caption{The singularity surfaces for the eight actuation modes}
      \label{singularity_all}
      \end{center}
   \end{minipage} \hfill
   \begin{minipage}[c]{.49\linewidth}
      \begin{center}
      \includegraphics[width=0.6\textwidth]{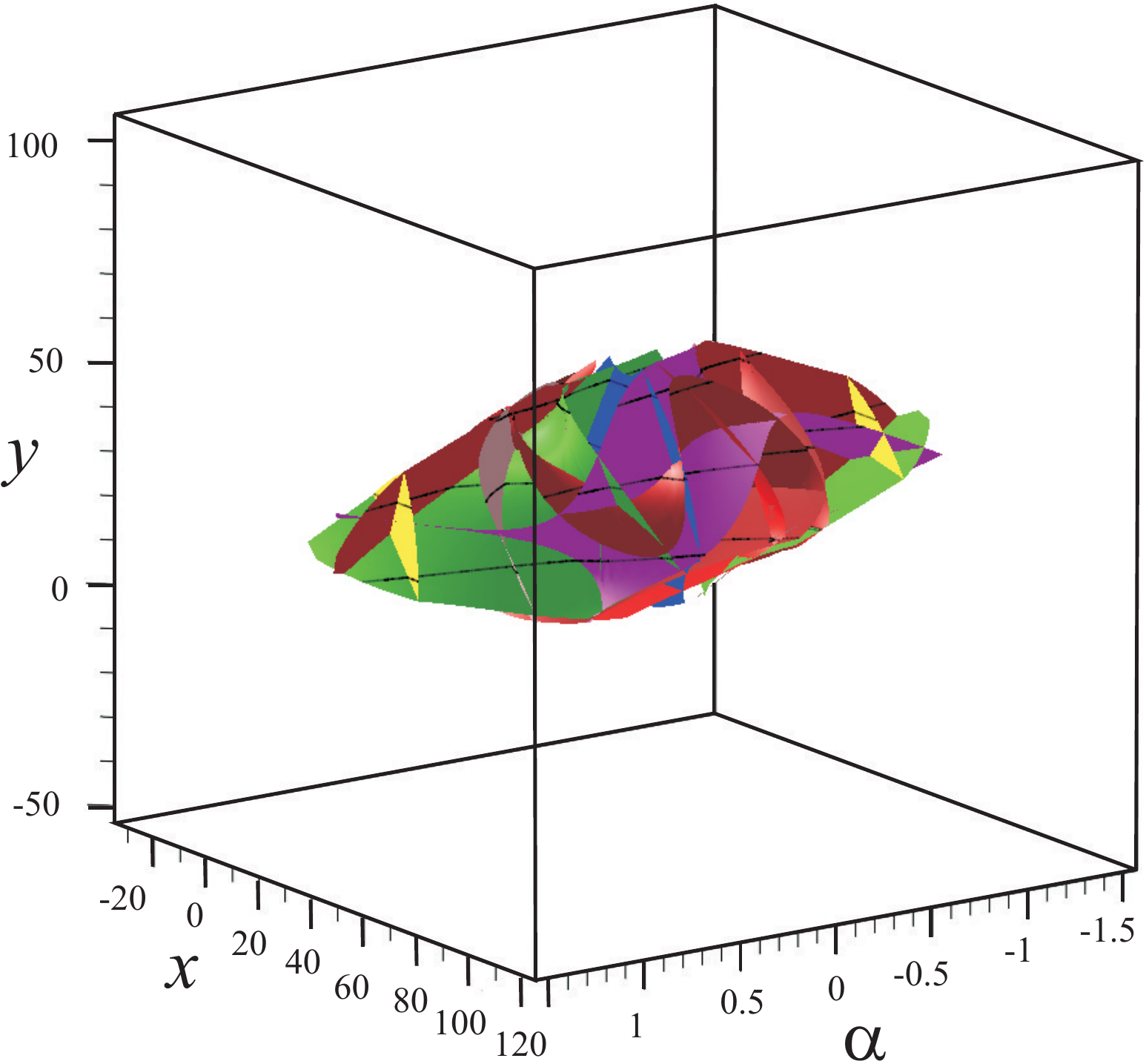}
      \caption{The singularity surfaces for the eight actuation modes within the robot workspace}
      \label{singularity_all_cut}
      \end{center}
   \end{minipage}
\end{figure}

Figures~\ref{singularity_all} and \ref{singularity_all_cut} show all singularities for the eight actuation modes without and with the joint limit conditions respectively. As none of them are superposed, it is possible to completely move through the workspace by choosing a non singular actuation mode for any pose of the mobile platform. The identical trajectory planning algorithm, which was introduced for the NaVARo I manipulator, can be used to select the actuation mode able to avoid singular configurations \cite{Caro:2014}. 

\begin{figure}
\begin{center}
\includegraphics[width=0.6\textwidth]{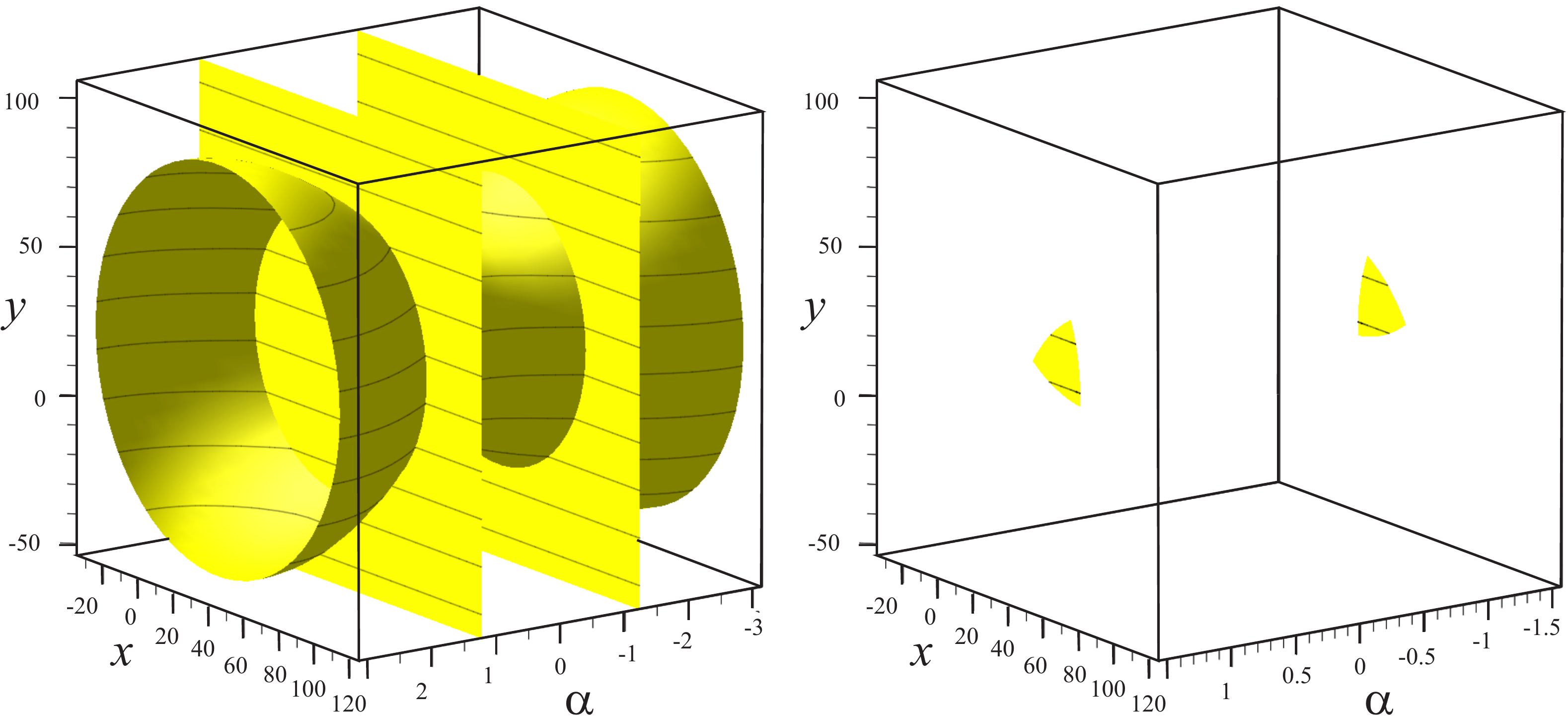}
\caption{Singularity surfaces for actuation modes 1}
\label{singularity_1}
\end{center}
\end{figure}
As an example, Figures~\ref{singularity_1} represents the singularities of the first actuation mode with, on the left, the singularities without joint limits and, on the right, those included in the workspace. 
\section{Conclusions}
In this article, a new version of the NaVARo manipulator was introduced and was derived from the well-known 3-RPR or even 3-RRR manipulators. Taking advantage of the actuation mode change, the entire Cartesian workspace can be used. By eliminating the parallelograms that allowed the first NaVARO robot to have an actuation on the second pivot joint which has the disadvantage of carrying mass, the Cartesian workspace is significantly increased. In addition, it is possible to install sensors on both actuated joints of each kinematic chain and to easily locate the mobile platform end-effector by solving the direct kinematic problem. The use of scissor mechanisms makes it possible to have a greater rigidity in the transverse direction of the robot movement as well as a prismatic-like displacement which can be increased largely according to the number of scissors. Unlike the NaVARo I, which is based on a 3-RRR robot, the NaVARo II  is based on the architecture of the 3-RPR, which has only one solution with the inverse kinematic model for any actuation mode. This property allows a complete writing of singularity equations whereas for the robot 3-RPR. In the literature, for the 3-RRR, these equations can only be written for a given orientation of the mobile platform. 
Future works will be carried out to evaluate the stiffness of the robot based on the size and the number of scissors and the number of solutions to the direct kinematic model to determine if there are actuation modes for which the robot is cuspidal. This problem can be expanded into an optimization problem where the scissor mechanisms will be tailored to the workspace requirements. 
\bibliographystyle{asmems4}

\section{Appendix}
\begin{figure*}[!hb]
{
\begin{eqnarray}
{\rm Actuation~mode~1~~~~~~~~~~~~~~~~~~~~~~~~~~~~~~~~~~~~~~~~~~~~~~~~~~~~~~~~~~~~~~~~~~~~~~~~~~~~~~~~~~~~~~~~~~~~~~~~~~~~~~~~~~~~~~~~~~~~~~~~~~~~~~~~~~}&& \nonumber \\
-3  (-30 \sqrt{3}y+x^2+y^2+1800 C_{\alpha} -90 x-300)  (C_{\alpha} -1/3) &=&0\\
{\rm Actuation~mode~2~~~~~~~~~~~~~~~~~~~~~~~~~~~~~~~~~~~~~~~~~~~~~~~~~~~~~~~~~~~~~~~~~~~~~~~~~~~~~~~~~~~~~~~~~~~~~~~~~~~~~~~~~~~~~~~~~~~~~~~~~~~~~~~~~~}&& \nonumber \\
( (-240 x+5400) y C_{\alpha}^2+ ( (120 x^2-120 y^2-5400 x) S_{\alpha} +360 y  (x-50)) C_{\alpha} &+&\nonumber\\
  (360 y^2+1800 x) S_{\alpha} +120 y  (x+45) ) \sqrt{3}+  (-120 x^2+120 y^2+16200 x) C_{\alpha}^2&+&\nonumber\\
  (( -240 x+16200) yS_{\alpha} +  (-4 x+180) y^2&-& \nonumber\\
4 x^{3}+540 x^2-18000 x) C_{\alpha} -4 y  (x^2+y^2-90 x+8550) S_{\alpha} -60 x^2-180 y^2-5400 x&=&0 \\
{\rm Actuation~mode~3~~~~~~~~~~~~~~~~~~~~~~~~~~~~~~~~~~~~~~~~~~~~~~~~~~~~~~~~~~~~~~~~~~~~~~~~~~~~~~~~~~~~~~~~~~~~~~~~~~~~~~~~~~~~~~~~~~~~~~~~~~~~~~~~~~}&& \nonumber \\
(-160 y  (x+45) C_{\alpha}^2+ ( (80 x^2-80 y^2+7200 x-648000) S_{\alpha} -4 y  (x^2+y^2-180 x-600) ) C_{\alpha} &+&\nonumber\\
  (4 x^{3}+4 xy^2-720 x^2+30000 x+108000) S_{\alpha} +80 y  (x-90) ) \sqrt{3}&+&\nonumber \\
  (-240 x^2+240 y^2+21600 x) C_{\alpha}^2+  (-480 y  (x-45) S_{\alpha} -4  (x-90)  (x^2+y^2-180 x-600) ) C_{\alpha} &-& \nonumber\\
4 y  (x^2+y^2-180 x+7500) S_{\alpha} +120 x^2-120 y^2-21600 x+972000&=&0\\
{\rm Actuation~mode~4~~~~~~~~~~~~~~~~~~~~~~~~~~~~~~~~~~~~~~~~~~~~~~~~~~~~~~~~~~~~~~~~~~~~~~~~~~~~~~~~~~~~~~~~~~~~~~~~~~~~~~~~~~~~~~~~~~~~~~~~~~~~~~~~~~}&& \nonumber \\
  (-160 y  (x-135) C_{\alpha}^2+  (( 80 x^2-80 y^2-21600 x+648000) S_{\alpha} +4 x^2y+4 y^{3}-34800 y) C_{\alpha} &+& \nonumber\\
  (-4 x^{3}+360 x^2+  (-4 y^2+2400) x+360 y^2-108000) S_{\alpha} +80 xy) \sqrt{3}+  (240 x^2-240 y^2-21600 x) C_{\alpha}^2&+&\nonumber\\ 
(480 y  (x-45) S_{\alpha} -4 x^{3}-4 xy^2+34800 x) C_{\alpha} +  (-4 x^2y-4 y^{3}+2400 y) S_{\alpha} -120 x^2+120 y^2&=&0\\
{\rm Actuation~mode~5~~~~~~~~~~~~~~~~~~~~~~~~~~~~~~~~~~~~~~~~~~~~~~~~~~~~~~~~~~~~~~~~~~~~~~~~~~~~~~~~~~~~~~~~~~~~~~~~~~~~~~~~~~~~~~~~~~~~~~~~~~~~~~~~~~}&& \nonumber \\
 ( (1680 xy-70200 y) C_{\alpha}^2+  (( -840 x^2+840 y^2+70200 x) S_{\alpha} -12 x^2y-12 y^{3}-2100 y) C_{\alpha} &+& \nonumber\\
  (12 x^{3}+12 xy^2-95100 x) S_{\alpha} -840 xy+32400 y) \sqrt{3}+  (840 x^2-840 y^2-210600 x+6804000) C_{\alpha}^2&+&\nonumber\\
  (( 1680 xy-210600 y) S_{\alpha} -36 x^{3}+3240 x^2+  (-36 y^2-6300) x+3240 y^2+1323000) C_{\alpha} &-& \nonumber \\
	36 y  (x^2+y^2-7925) S_{\alpha} + 180 x^2+1020 y^2+97200 x-6902000&=&0\\
{\rm Actuation~mode~6~~~~~~~~~~~~~~~~~~~~~~~~~~~~~~~~~~~~~~~~~~~~~~~~~~~~~~~~~~~~~~~~~~~~~~~~~~~~~~~~~~~~~~~~~~~~~~~~~~~~~~~~~~~~~~~~~~~~~~~~~~~~~~~~~~}&& \nonumber \\
( (80 x^2-80 y^2-7200 x+324000) C_{\alpha}^2+  (160 y  (x-45) S_{\alpha} +360 y^2-54000) C_{\alpha} &-& \nonumber\\
360 y  (x-45) S_{\alpha} -60 x^2+20 y^2+5400 x-156000) \sqrt{3}&-&\nonumber\\
4 y  (x^2+y^2-90 x+8100) C_{\alpha} +4  (x-45)  (x^2+y^2-90 x) S_{\alpha} +5400 y&=&0\\
 {\rm Actuation~mode~7~~~~~~~~~~~~~~~~~~~~~~~~~~~~~~~~~~~~~~~~~~~~~~~~~~~~~~~~~~~~~~~~~~~~~~~~~~~~~~~~~~~~~~~~~~~~~~~~~~~~~~~~~~~~~~~~~~~~~~~~~~~~~~~~~~}&& \nonumber \\
(480 y (x-45) C_{\alpha}^2+ ( (-240 x^2+240 y^2+21600 x) S_{\alpha} +4 y (x^2+y^2-180 x+8100) ) C_{\alpha} &-& \nonumber\\
4 (x-90) (x^2+y^2-180 x) S_{\alpha} -240 y (x-45))\sqrt{3}+(1944000-240 x^2+240 y^2-21600 x) C_{\alpha}^2&+&\nonumber\\
 (-480 y  (x+45)S_{\alpha} -324000-12x^{3}+2160 x^2+ (-12 y^2-97200) x) C_{\alpha} -12 y  (x^2+y^2-180 x) S_{\alpha} &-& \nonumber \\
240 y^2+32400 x-936000&=&0\\
{\rm Actuation~mode~8~~~~~~~~~~~~~~~~~~~~~~~~~~~~~~~~~~~~~~~~~~~~~~~~~~~~~~~~~~~~~~~~~~~~~~~~~~~~~~~~~~~~~~~~~~~~~~~~~~~~~~~~~~~~~~~~~~~~~~~~~~~~~~~~~~}&& \nonumber \\
-S_{\alpha}  (30 \sqrt{3}y-x^2-y^2-1800 C_{\alpha} +90 x+300) &=&0 
\end{eqnarray}}
\end{figure*}
\end{document}